\documentclass[letterpaper, 10 pt, conference]{ieeeconf}  

\IEEEoverridecommandlockouts                              
\usepackage{times}
\usepackage{helvet}
\usepackage{courier}

\usepackage{color}
\usepackage{amsmath}
\usepackage{amssymb}
\usepackage[ruled,vlined,linesnumbered]{algorithm2e}
\usepackage{graphicx}
\usepackage{epsfig}

\newtheorem{theorem}{Theorem}
\newtheorem{lemma}{Lemma}

\title{\LARGE \bf Simultaneous Configuration Formation and Information Collection by Modular Robotic Systems}
\author{Ayan Dutta, Prithviraj Dasgupta
\thanks{Computer Science Department, University of Nebraska at Omaha, USA.
{\tt\small \{adutta, pdasgupta\}@unomaha.edu}}%
\thanks{*The authors would like to acknowledge the contributions of Prof. Carl Nelson, Dept. of Mechanical and Materials Engg., Univ. of Nebraska-Lincoln in designing the robot hardware in the ModRED project.}
}

\long\def\omitit#1{}

\begin{document}
\maketitle
\thispagestyle{empty}
\pagestyle{empty}

\begin{abstract}
We  consider the configuration formation problem in modular robotic systems where a set of singleton modules that are spatially distributed in an environment are required to assume appropriate positions so that they can configure into a new, user-specified target configuration, while simultaneously maximizing the amount of information collected while navigating from their initial to final positions. Each module has a limited energy budget to expend while moving from its initial to goal location. To solve this problem, we propose a budget-limited, heuristic search-based algorithm that finds a path that maximizes the entropy of the expected information along the path. We have analytically proved that our proposed approach converges within finite time. Experimental results show that our planning approach has lower run-time than an auction-based allocation algorithm for selecting modules' spots. 
\end{abstract}


\section{Introduction}
Modular self-reconfigurable robots (MSRs) \cite{Stoy10} are composed of individual robotic modules which can change their connections with each other to form different shapes or configurations. MSRs are highly dexterous and maneuverable as they can change their shape or configuration to adapt to the environment or task at hand. A central problem in MSRs is the configuration formation problem - given a set of modules initially distributed arbitrarily within the environment and a desired target configuration involving those modules, how can each module select an appropriate spot or location in the target configuration to move to, so that, after reaching the position, it can readily connect with adjacent modules and form the shape of the desired target configuration. As a motivating example, we consider a scenario where a set of singleton modules are collecting information from an environment. To access a specific region of the environment, e.g., an elevated region, they need to form a certain shape (configuration) such as a legged configuration, which allows them to navigate the elevation. To get into this new shape, all the singleton modules will plan their paths from their current locations to appropriate positions in the target configuration. We consider the additional navigation criterion for information collection - the modules have to select their navigation paths so that they can increase the amount of information they collect such as soil/rock sample collection, temperature measurement etc. using their on-board sensors, while they are moving towards their positions in the target configuration.

The configuration formation problem is challenging as the most preferred position (e.g., position involving least time and battery expenditure to navigate to) of one module in the target configuration could conflict with the most preferred position of another module, leading to possible deadlocks that could result in failed attempts to achieve the target configuration. Simultaneously, it is non-trivial to plan a module's path to maximize the information it can collect because the distribution of information in the environment is not known {\em a priori} and a brute-force exploration might unnecessarily expend the limited battery budget on the module. In this paper, we combine these two problems into a single problem called the {\em Simultaneous Configuration Formation and Information Collection} and solve the problem by using a heuristics search-based algorithm along with an entropy-maximization-based, dynamic path planning approach.
We have proved that modules using our proposed approach can form the given target configuration in finite time. Our experimental results show that our proposed planning strategy outperforms a comparable, auction based allocation mechanism in terms of run-time and number of messages exchanged.

\begin{figure}
\begin{center}
\includegraphics[width=\linewidth]{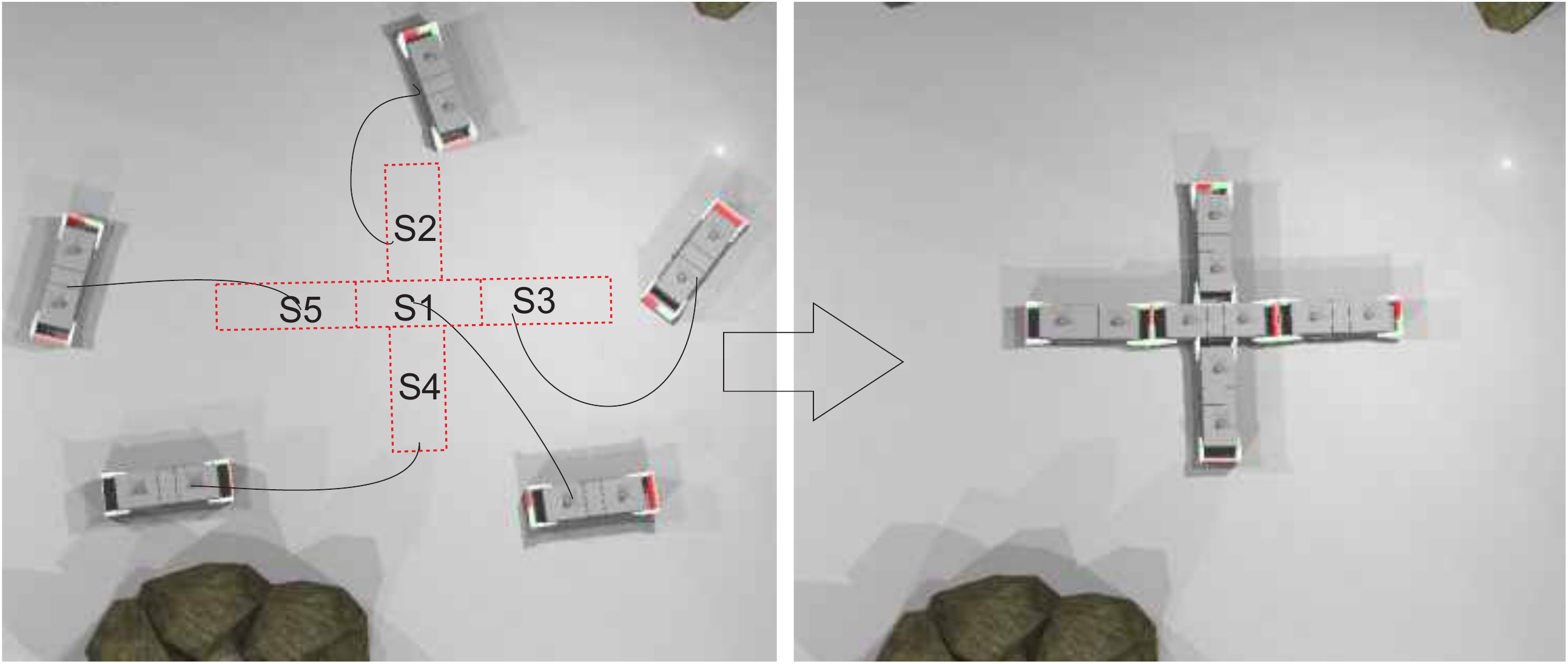}
\end{center}
\caption{Illustrative example of the problem: randomly distributed ModRED modules \cite{baca2014modred} follow maximally possible informative paths to assume the goal spots in the target configuration (left) and finally the configuration is formed (right).}
\label{illustration}
\end{figure}

\section{Related Work}
{\bf Configuration Formation: }An excellent overview of the state of the art MSRs and related techniques is given in \cite{Stoy10}. Configuration formation is a way to fulfill shape-formation function, in which modules aggregate autonomously to a final shape or configuration. In the context of MSRs, configuration formation enables modular robots to transform into any desired configuration. Configuration formation in modular robot systems have been studied less extensively \cite{ahmadzadeh2015modular}. A few studies on configuration formation (by means of programmable self-assembly) can be found for self-actuated modular robots \cite{klavins2007programmable}, and for modules that lack innate actuation
ability, like stochastically-driven modules in liquid environment \cite{tolley2010fluidic}. 
In swarm robotic systems, there are many studies on autonomous self-assembly of robot swarms. Alonso-Mora {\em et al.}~\cite{alonso2011multi} have addressed the problem of artistic pattern formation by robot swarms where robots are initially distributed arbitrarily (spatially) in an environment and are required to assemble to form a certain pattern. Similarly, distributed algorithms for robotic construction~\cite{werfel2008three} have been proposed to solve the problem of allocating blocks mobilized by robots to certain positions in a target configuration. Specific pattern, such as circle formation by asynchronous robots has been studied in \cite{dutta2012circle, datta2013circle}.

{\bf Informative Path Planning: }Mutli-robot informative path planning (MIPP) involves an aspect of the general multi-robot path planning problem where each robot has to determine waypoints between given start and end locations in the environment so that the information gain of the resulting path is increased. In one the earliest works on MIPP, Singh {\em et al.} \cite{Singh09} have proposed a recursive, branch and bound algorithm to solve the MIPP problem that finds the best, budget-limited path through a graph of possible way-points. 
The MIPP problem with periodic connectivity between robots has been studied in \cite{Hollinger10}. 
In \cite{hollinger2013sampling}, authors have proposed a sampling based technique for information collection. 
In \cite{hitz2014fully}, authors have proposed a dynamic programming based approach for autonomous monitoring in an environment modeled as a transect. To the best of our knowledge, our work is one of the first attempts to merge these two problems where a set of initial singleton modules need to form a certain configuration while maximizing the amount of information collected on the way to forming that configuration. 

\section{Problem Setup}
Let $\mathbb{A}=\{a_1, a_2,...\}$ denote a set of robot modules. Each $a_i \in \mathbb{A}$ has an initial pose denoted by $a_i^{pos} = (x_i, y_i, \theta_i)$, where $(x_i, y_i)$ denotes the location of $a_i$ and $\theta_i$ denotes its orientation within a $2$D plane corresponding to the environment. Each module has a unique identifier. For the purpose of navigation, each module uses a map of the environment; the map is decomposed into grid-like cells using a cellular decomposition technique. We assume that initially all the modules are within each others' communication range.

In the variant of configuration formation problem studied in this paper, singleton robot modules, starting from arbitrary initial locations, are required to get into a specified target configuration. The target configuration is represented as a graph, denoted by $G_T=(V_T,E_T)$, where $V_T= \{s_1, s_2,...\}$ is the set of vertices and $E_T=\{e_{ij}=(s_i, s_j)\}$ is the set of edges. Each vertex in $V_T$ is referred to as a {\em spot} that a module needs to occupy. Each spot $s_i \in V_T$ is specified by its pose and its neighboring spots in the target configuration, $s_i = (s_i^{pos}, neigh(s_i))$, where $neigh(s_i) \subset V_T$.

For information collection purpose a robot needs to sense the region it is situated in with its sensors. We discretize the information collection procedure, by using $\mathcal{C}$ to denote the set of information collection locations or cells in the environment. $\mathcal{C}$ can be decomposed into two disjoint subsets, $O$ and $U$, corresponding to the cells that are visited and not visited by the robots. Note that, $V_T \subset \mathcal{C}$. Robot $a_i$'s path from its current location to a spot, $s_j$, in a target configuration is defined as an ordered sequence of cells it visits, i.e., $P_{ij} = \{c_{1}, c_{2}, ..., s_j\} \subseteq \mathcal{C}$. Cost of a path $P_{ij}$ is defined by the number of cells present in that path, i.e., $cost(P_{ij}) = |P_{ij}| - 1$.

To model the environmental phenomena generating the information, we have used Gaussian processes (GP)~\cite{Guestrin05, hitz2014fully}. Modeling the environment as a GP requires the assumption that all the sampling locations $\mathcal{C}$ in the environment have a joint Gaussian distribution. A GP can be defined by its mean $m(\cdot)$ and its co-variance (\textit{kernel}) $k(\cdot,\cdot)$ functions. Given a set of measurements $X_O$, we can predict the information measurement in the rest of the unobserved locations $U$, conditioned on $X_O$. A GP can be specified by the following equations \cite{Guestrin05}:
\[\mu_{U|O} = \mu_U + \Sigma_{UO}\Sigma_{OO}^{-1} (X_O - \mu_O)\]

\[
\sigma^2_{U|O} = \Sigma_{UU} - \Sigma_{UO}\Sigma_{OO}^{-1}\Sigma_{OU}
\]
where $\mu_{U|O}$ is the conditional mean and $\sigma^2_{U|O}$ is the variance. $\Sigma_{UO}$ is the co-variance matrix, with an entry for every location $o \in O$. Following GP formulations, the objective of informative path planning is to plan a path which maximizes the entropy, where entropy is given by:
\begin{equation}
\label{eq_entropy}
H(U|O) = \frac{1}{2}log (2 \pi e \sigma^2_{U|O})
\end{equation}
The main idea behind entropy maximization is to select the locations in the environment, which have the highest amount of uncertainty.

We have modeled the path planning with information collection problem as an instance of the bounded-cost search problem~\cite{stern2014potential}. In this problem, the evaluation function for a cell is called its {\em potential}. The potential of a cell $c$ is defined in our problem as $u(c) = \frac{B - g(c)}{h(c)}$, where $g(c)$ is the cost of moving from the start cell(location) to cell(location) $c$, $h(c)$ is the estimated cost of moving from cell(location) $c$ to the goal location, and, $B$, is the budget that corresponds to the maximum of number of cells in any module's path from its current position to the goal location, i.e., maximum allowable path length. From this it follows that the cost of the path used by module $a_i$ to occupy spot $s_j$ is budget-limited to $B$, i.e., $cost(P_{ij}) \le B, \forall a_i \in \mathbb{A}, s_j \in V_T$. The informativeness of path $P_{ij}$ is computed as $inf(P_{ij}) = \sum \limits_{\forall c_k \in P_{ij}} H(c_k)$. 

For finding the path from every module's current location to its goal position in the target configuration, a best-first technique is used which explores nodes with larger {\em entropic potential} ($hu(\cdot)$) values, defined as $hu(c) = u(c) + H(c|O)$. Formally we can define the studied problem as follows: Given a set of singleton modules $\mathbb{A}$ and a set of spots $V_T$ representing the target configuration, find a suitable allocation $f : \mathbb{A} \rightarrow V_T$ such that $\forall a_i \in \mathbb{A}, \quad P_{ij} = \displaystyle \arg \max_{P \in \Pi, s_j \in V_T} inf(P)$ and $cost(P_{ij}) < B$; $\forall a_k \neq a_i, \quad f(a_i) \neq f(a_k)$, where $\Pi$ denotes the set of all possible paths from $a_i$'s current location to the goal location.

\section{Algorithm Description}
The solution approach is divided into two phases - a {\it planning phase}, where modules select spots in the target configuration and an {\it acting phase}, where modules move to their selected spots.

\subsection{Planning Phase}
In the beginning of the planning phase, all the modules broadcast their positions and orientations. We assume that each module autonomously and independently plans its paths to all the spots, and a module is aware of only its local planning information for any spot. Consequently, multiple modules could have identical maximum informative paths for the same spot and end up choosing it to move to. This could result in occlusions to each other, and, in the worst case, a failure of the configuration process. To avoid such a situation, we propose an additional coordination mechanism by employing a centralized supervisor to resolve conflicts between modules for the same spots in a structured manner, without incurring a high computational overhead.

\textbf{Computing Informative Paths using Entropic Potential Search (EPS) Algorithm}: Our proposed planning mechanism operates in two phases, as shown in Algorithm \ref{spot_algo}. In the first phase, called the \textit{computation} phase, each module $a_i$ first calculates informativeness of the paths from its current location to each of the spots in $V_T$, using the Entropic Potential Search algorithm (EPS) (Algorithm \ref{EPS}). This is a modified version of the PTS algorithm proposed by Stern et al. \cite{stern2014potential}. The algorithm employs a greedy best-first technique to explore the cells with high entropic potential values. The EPS algorithm takes a module's current location and one of the positions in the target configuration as input, along with the bounded cost (budget) $B$. A data structure, called $OPEN$, is maintained for holding nodes for further exploration. Another data structure, called $CLOSED$, is maintained for holding the nodes which have been explored already.

In each iteration, the node, $n_{max}$, with the highest entropic potential value is expanded. If the current neighbor cell, $n_n$, of $n_{max}$ is already in $OPEN$ with smaller or equal $g(\cdot)$ value, then $n_n$ is ignored. Because we assume the heuristic function, $h(\cdot)$, to be admissible, it is necessary to check whether $g(n_n) + h(n_n)$ surpasses $B$. If $g(n_n) + h(n_n) > B$, then $n_n$ is pruned, as it can never be a part of the required bounded cost solution. If $n_n$ is the goal cell, then the search procedure terminates. Otherwise, $n_n$ is pushed back into $OPEN$, if the entropy value of cell $n_n$, $H(n_n)$, is greater than $0$, and the search continues{\footnote{Initial cells of the modules have been treated as obstacles and therefore restricted to be added to $OPEN$.}}. This way we never explore a cell which does not guarantee to have any entropy value. Once EPS is terminated either we find a path with cost lower than $B$ which is also highly informative or EPS returns null to notify that no such path with cost lower than $B$ exists. 

Every module individually runs the EPS algorithm for every spot $s_j \in V_T$. 
Each module sends its list of spots with computed informativeness to a supervisor node for the following \textit{allocation} phase.{\footnote{The supervisor could be a centralized external entity or one of the modules with higher computational capabilities elected using a leader election protocol.}}

\textbf{Allocation}: During the \textit{allocation} phase, the supervisor waits until it receives the sorted lists of spots from all the modules. Then it proceeds to allocate spots in rounds, while allocating one spot in each round, starting from $s_1$. In round $j$, spot $s_j$ is allocated to the module $a_i$ that has the highest informative path $inf(P_{ij})$ to $s_j$. If a module is allocated in a certain round, it is not considered for allocation in subsequent rounds. In case every available module's path cost exceeds budget $B$, it means that there is no module available that can occupy $s_j$ while remaining within the battery constraint. In such a case, the module that has the lowest cost path $P_{ij}$ among the conflicted modules for spot $s_j$ is allocated to $s_j$. A similar strategy is used even if all the modules have the same informative paths for a specific spot, where path cost is below $B$. If ties still remain after applying the above strategy, they are broken at random. At the end of the allocation phase, the supervisor sends the list of allocated spots to all the modules.

\begin{algorithm}
{\bf Phase 1: Computation Phase by Modules}\\
Each module $a_i$ will do the following:\\
For all spots $s_j \in V_T$, execute {\em pathFormation()} algorithm and find a set of paths, $P$, to all spots.\\
Send the list of spots along with the informativeness values of all paths to all spots to the supervisor

{\bf Phase 2: Spot Allocation by Supervisor}\\
wait until ranked list of slots recd. from all modules\\
\For {each spot $s_j$}
{
$winners \leftarrow \arg \max_{a_i \in \mathbb{A}} inf(P_{ij})$\\
\If {only one module $a_i$ in $winners$}
{
$winner \leftarrow a_i$
}
\Else 
{
// more than one winner module: multiple modules with same informativeness for $s_j$\\
$winner \leftarrow \arg \min_{a_i \in winners} cost(P_{ij})$;\\
// ties are broken randomly
}
add $(winner, s_j)$ to $f(\cdot)$;\\
remove $winner$ from $\mathbb{A}$ and remove $s_j$ from $V_T$;

}
Send set of spot allotments $f(\cdot)$ to every module $a_i$.
\caption{Spot Allocation (SA) Algorithm}
\label{spot_algo}
\end{algorithm}

\begin{algorithm}[ht!]
{\em pathFormation()}\\
\KwIn{$B$: Budgeted cost; $c_{curr}$: Current cell of the module and $s_{k}$: A node in the target configuration.}
\KwOut{$P_{ik} \in \Pi$: Generated path for module $r_i$.}
$OPEN \leftarrow c_{curr}$.\\
$CLOSED \leftarrow \{\emptyset\}$.\\
\While{OPEN is not empty}{
$n_{max} \leftarrow$ arg $\max \limits_{n \in OPEN} hu(n)$\\
	\For{each neighbor $n_n$ of n}{
	
	\If{$n_n$ is in $OPEN$ or $CLOSED$ and $g(n_n) \le g(n_{max}) + 	cost(n_{max}, n_n)$}{
			Continue with the next neighbor of $n$.
	}
	$g(n_n) \leftarrow g(n_{max}) + 	cost(n_{max}, n_n)$\\
	\If {$g(n_n) + h(n_n) \ge B$}{
	Continue to the next successor of $n_{max}$}

	\If{$n_n = s_k$}{
	 return the best path to $s_k \rightarrow P_{ik}$}
	\eIf{$n_n \in OPEN$}{
	Update the $g(n_n)$ value of $n_n$ in $OPEN$}
	{	
		\If{$H(n_n) > 0$}{Insert $n_n$ to $OPEN$}
	}
	}
	Insert $n_{max}$ to $CLOSED$
}
return Null // no solution exists which has lower cost than $B$
\caption{Entropic Potential Search (EPS) Algorithm}
\label{EPS}
\end{algorithm}

\subsection{Acting Phase}
In the acting phase, the modules move to their respective allocated spots in a sequential manner. No module is allowed to move until all the spots are allocated using the allocation phase. In the absence of a proper order of modules to occupy spots, deadlock situations might arise. 
For example, in Figure \ref{illustration}, if all the spots except S$_1$ are assumed first, then the module which has selected the spot S$_1$ arrives, it will not be able to move to S$_1$, unless other modules disconnect and make space for it to move. To avoid repeated connects and disconnects between modules, we allow the module which has selected the spot with the highest betweenness centrality measure in $G_T$ \cite{brandes2001faster}, first to occupy its position (ties are broken at random). Once it is in its proper position, it will broadcast message to notify that it has concluded locomotion, to all other modules. Next the spots neighboring the center spot will be occupied by modules and so on. Techniques described in \cite{baca2013modular} can be used for locomotion purpose of the modules. 

Each module, $a_i$, maintains a list of its visited cells, $C_{V_i} \in \mathcal{C}$, while moving towards its goal position in the target configuration. In a GP, with newly added set of visited cells, the estimated entropy of the unobserved cells gets updated as given by Equation \ref{eq_entropy}. To incorporate this change and also to gain maximum information from the environment, modules need to update their paths, whenever possible. Modules update their initially calculated paths by following Algorithm \ref{act_algo}. After visiting $\mathcal{O}$ new cells, each module executes the EPS algorithm with its remaining budget.

\begin{algorithm}
\KwIn{$B_r$: Remaining budget; $c_{curr}$: Current cell of the module and $s_{k} \in V_T$: Goal position in the target configuration.}
$\bar{P}_{ik} \subset P_{ik}$: Module $a_i$'s remaining path from $c_{curr}$ to $s_{k}$.\\
Update the set of visited cells, $C_{V_i}$.\\
\If{module $a_i$ has visited $\mathcal{O}$ cells}{
	Execute {\em pathFormation($B_r, c_{curr}, s_{k}$)} to find a new path, $P_{ik}^*$.\\
	\eIf{$inf(P_{ik}^*) > inf(\bar{P}_{ik})$ and $cost(P_{ik}^*) \le B_r$}
		{Follow the new path $P_{ik}^*$\\
			$\bar{P}_{ik} \leftarrow P_{ik}^*$			
			}
	{Follow initially generated path $\bar{P}_{ik}$}
}
\If{module $a_i$ has reached its goal position $s_{k} \in V_T$}
{Broadcast REACHED message}
\caption{Movement Strategy Of Modules}
\label{act_algo}
\end{algorithm}

If a new path from the module's current cell to the goal position can be found while remaining within the budget constraint and  improving the informativeness, then the module selects it to move towards its allocated spot. Otherwise it follows the earlier path $\bar{P}_{ik}$. Once a module reaches its goal position in the target configuration, it broadcasts a REACHED message to notify other modules. Modules are allowed to move exclusively in the order of the centrality of selected spots; ties are broken at random.

\subsection{Theoretical Analysis}
\begin{lemma}
Final formed configuration will contain no hole, if $|\mathbb{A}| \ge |S|$.
\end{lemma}
{\it Proof}: We prove this by contradiction. Let's assume that there is a hole in the final configuration, i.e., a spot $s_{h}$ is not assumed by any module. This can happen because either of the two reasons: $1$. No module has selected $s_{h}$, or $2$. module $a_{h}$, which selected $s_{h}$, could not reach that spot because other modules blocked its way, to its selected spot. These two situations cannot arise. If $|\mathbb{A}| \ge |S|$, then supervisor will allocate each spot to a unique module. 
So we can guarantee that some $a_h$ will select $s_h$. Secondly, from our model of sequential module movement (acting phase), we can guarantee that at first the spots nearer to \textit{center} are assumed and then the outer ones. So, no outer spot will be filled, before its neighbor, nearer to the center got filled. Hence, we can guarantee that there will be no hole in the final formed configuration.

\begin{theorem}
Algorithms \ref{spot_algo} and \ref{act_algo} will eventually converge and modules will form the desired target configuration. 
\end{theorem}
{\it Proof}: In lemma $1$, we have proved that there will be no hole in the target configuration. And as the modules have finite speed of locomotion, we can say that eventually the target configuration will be formed.



\textbf{Complexity Analysis}: 
The worst case time complexity of the EPS algorithm is $O(b^d)$, where $b$ is the branching factor of cell $c \in \mathcal{C}$ and $d$ is the maximum length of the solution. For a $4$-connected environment and for given budget $B$, complexity becomes $O(4^B)$. Each module runs the EPS algorithm for every spot - making the worst case time complexity for each module $O(|V_T| \cdot 4^B)$. In the acting phase, in the worst case scenario, any module might run the EPS algorithm $\mathcal{Z}(=\frac{B}{\mathcal{O}})$ times, which makes the worst case time complexity for each module to be $O( 4^B \cdot (|V_T| + \mathcal{Z}))$. Worst case time complexity for the supervisor is $O(|\mathbb{A}| \cdot |V_T|)$.

\begin{figure}[ht!]
\begin{center}
\begin{tabular}{ccc}
\hspace{-0.1in}\includegraphics[width=0.32\linewidth]{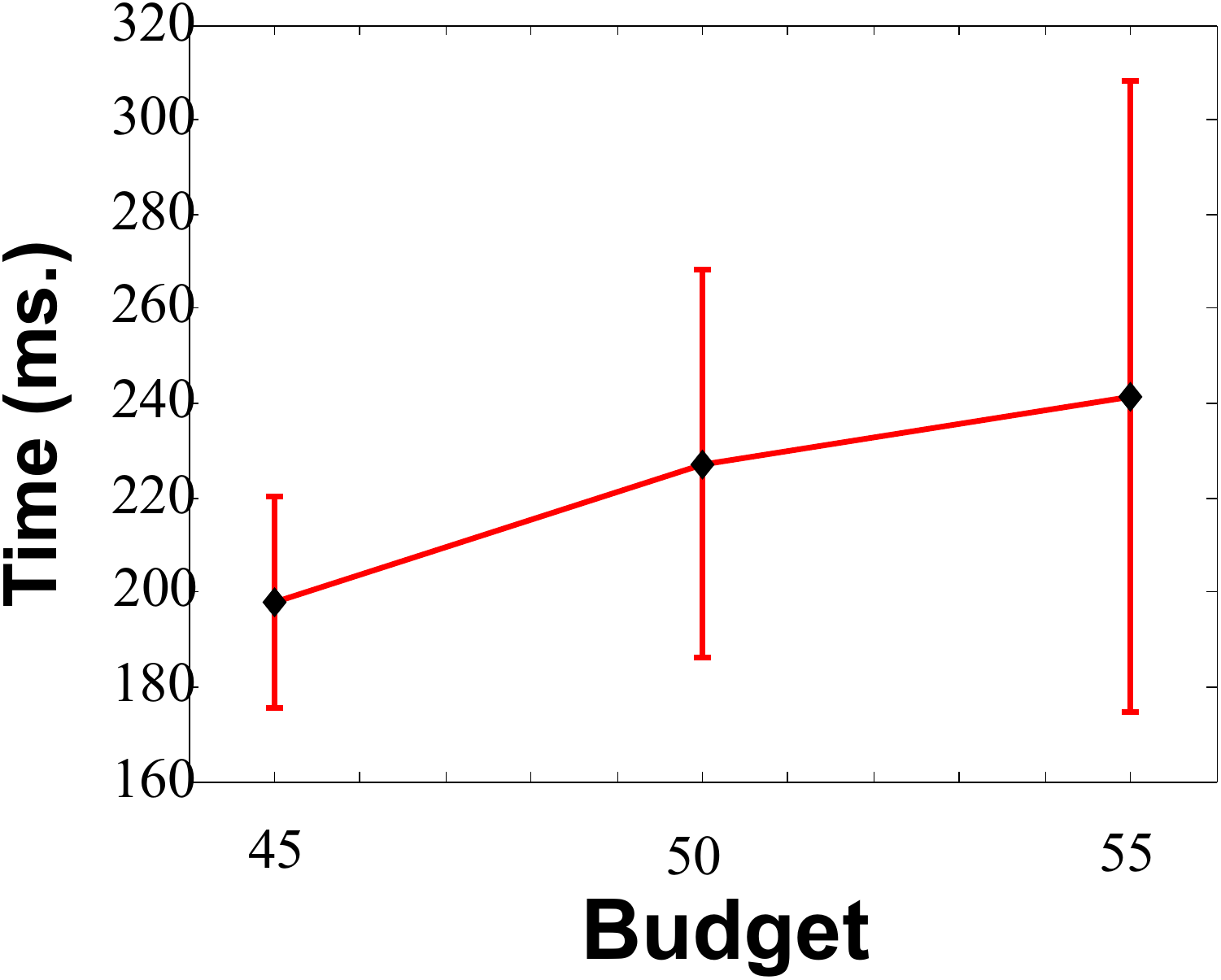}&
\hspace{-0.1in}\includegraphics[width=0.33\linewidth]{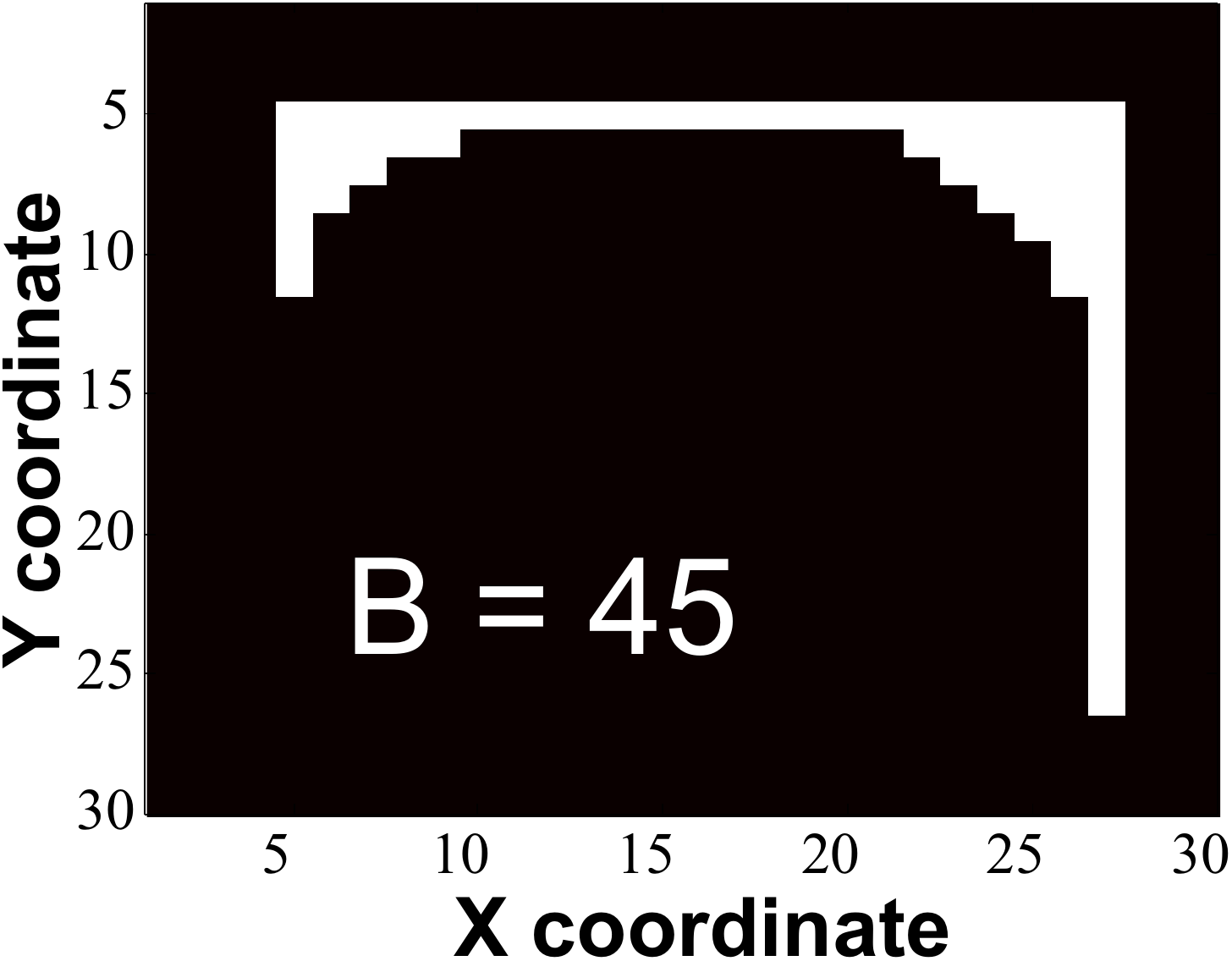}&
\hspace{-0.1in}\includegraphics[width=0.33\linewidth]{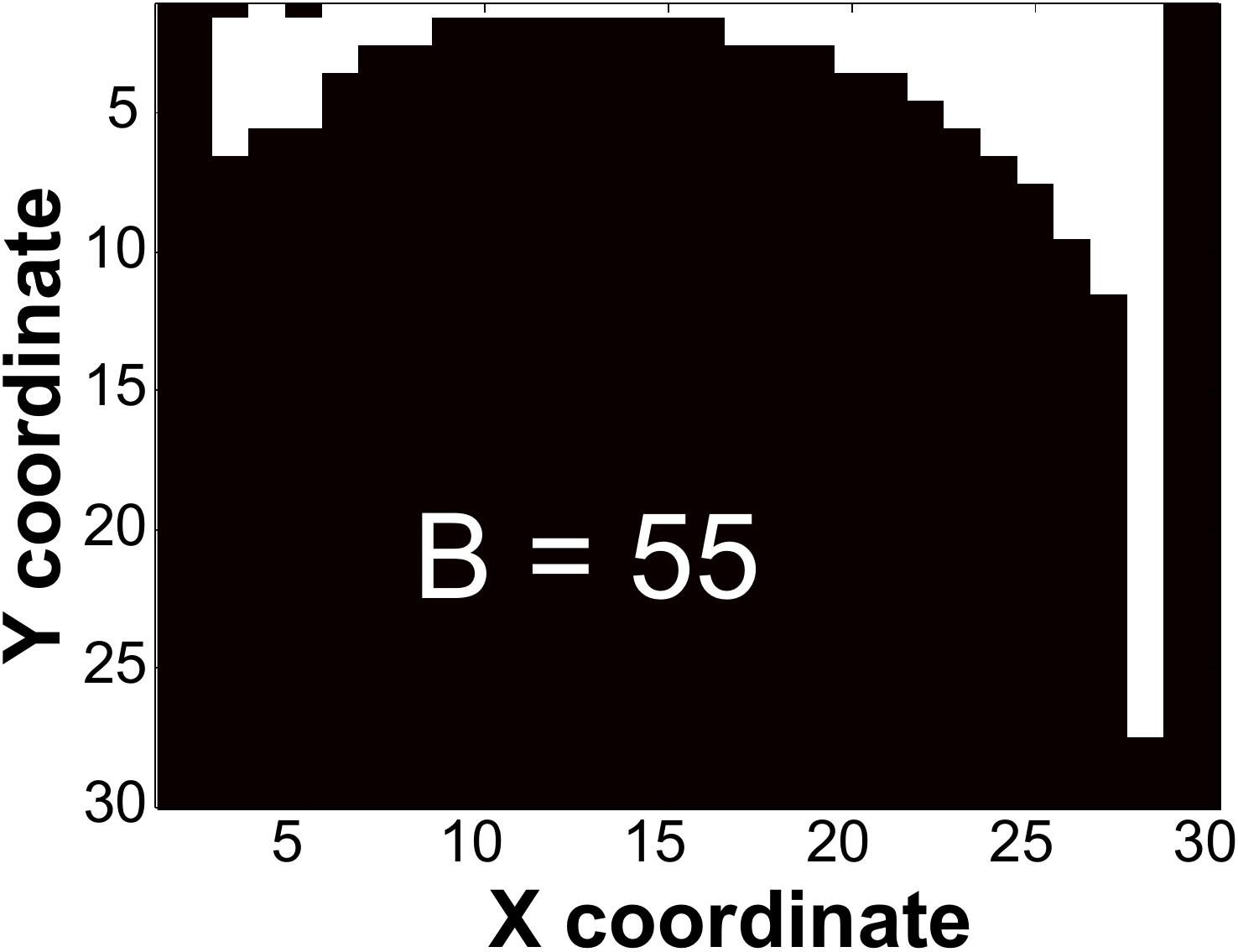}\\
(a) & (b) & (c)\\
\end{tabular}
\end{center}
\caption{(a) Run times of EPS algorithm for different budgets; (b), (c) Nodes explored (shown in white color) by EPS algorithm, with $B = 45$ and $55$ respectively.}
\label{timetable}
\end{figure}

\section{Experimental Evaluation}
{\bf Experimental Settings.} We have implemented the algorithms in simulation on a desktop PC (Intel Core i5 -960 3.20GHz, 6GB DDR3 SDRAM). The environment is divided into a $30 \times 30, 4$-connected grid structure. Each cell in the environment is represented by its centroid. The information value of each cell in the environment is drawn from $U[1,10]$. We have tested instances where random target configurations, in forms of graphs, have been generated of sizes, $\mid V_T \mid = 10$ through $50$, inside the environment. Each node in the target configuration has between $1$ to $4$ neighbor nodes and each edge between two neighbor nodes has unit distance. In all the cases, $|\mathbb{A}| = |V_T|$. Each module is modeled to be a cube of size $1$ unit $\times 1$ unit $\times 1$ unit; their initial cells are drawn uniformly from $\mathbb{U}[(0, 29), (0,29)]$. Similar to \cite{low2012decentralized}, $\frac{2}{5}$-th of total cells and their corresponding ground truth data has been provided to the modules to learn the mean and covariance structure of GP through maximum likelihood estimation. Budget, $B$, has been set to $45$ cells unless otherwise mentioned. We have used Manhattan Distance ($\mathcal{MD}$) for calculating cost of a path. Each singleton module runs the SA algorithm and then moves to its allocated or selected spot in $G_T$. Each test is run $5$ times. 

We have also compared the performance of the SA algorithm with an auction algorithm \cite{bertsekas1990auction}, which is a classical assignment algorithm. For implementing auction algorithm, each module is modeled as a bidder and each spot is modeled as an item, which modules are bidding for. 

\begin{figure*}[ht!]
\begin{center}
\begin{tabular}{cccc}
\hspace{-0.3in}\includegraphics[width=0.23\linewidth]{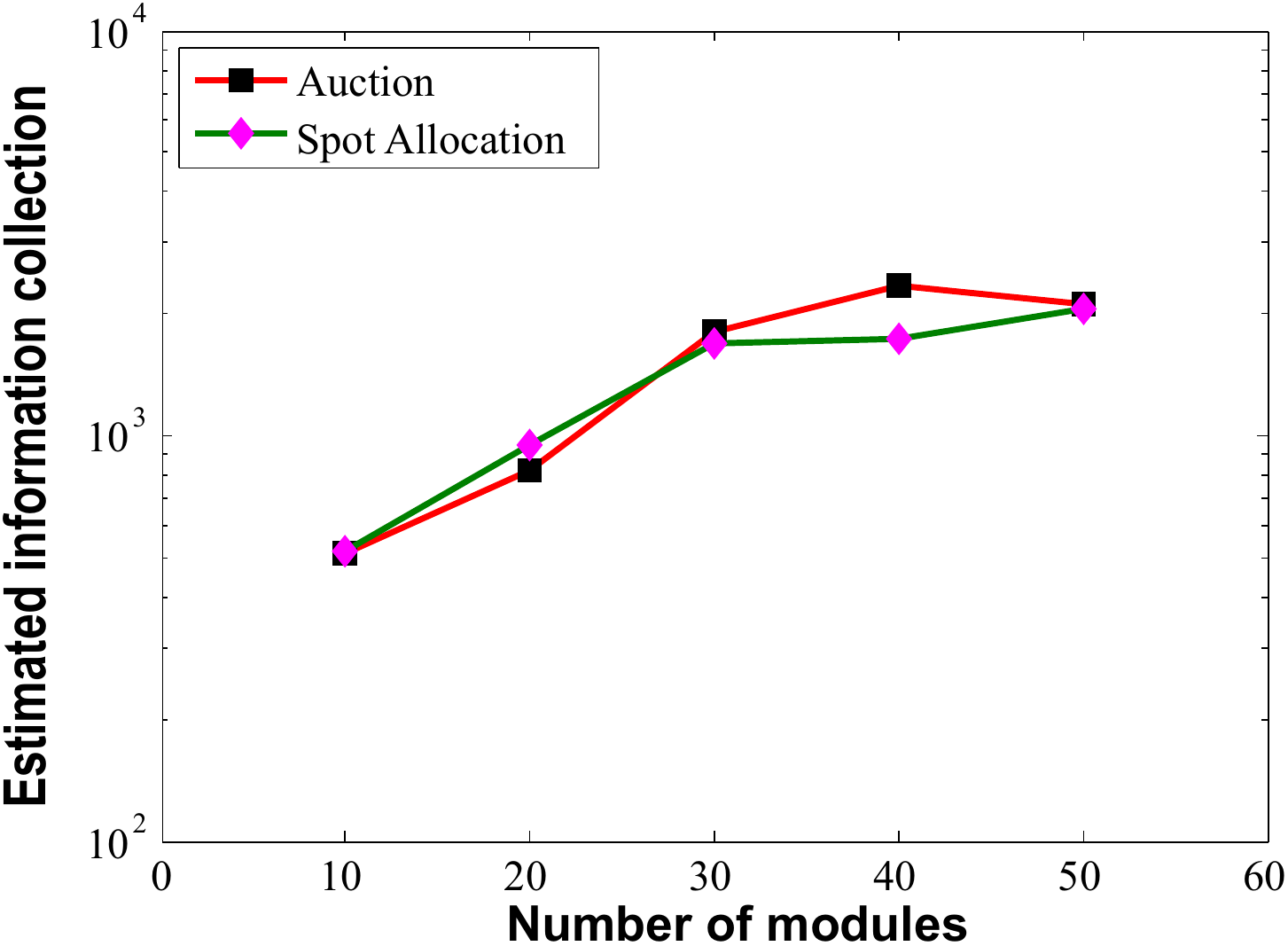}&
\hspace{-0.1in}\includegraphics[width=0.23\linewidth]{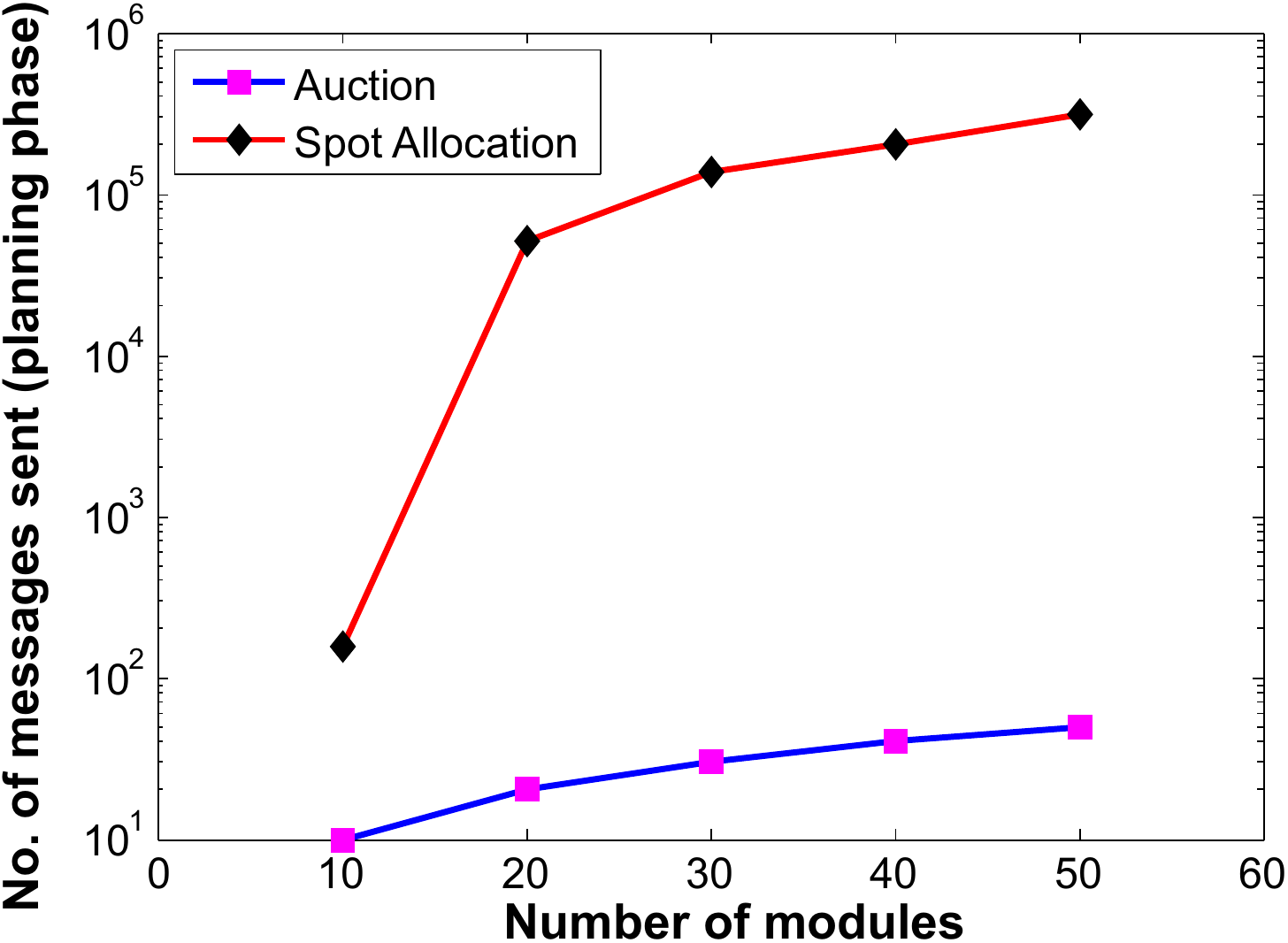}&
\hspace{-0.1in}\includegraphics[width=0.23\linewidth]{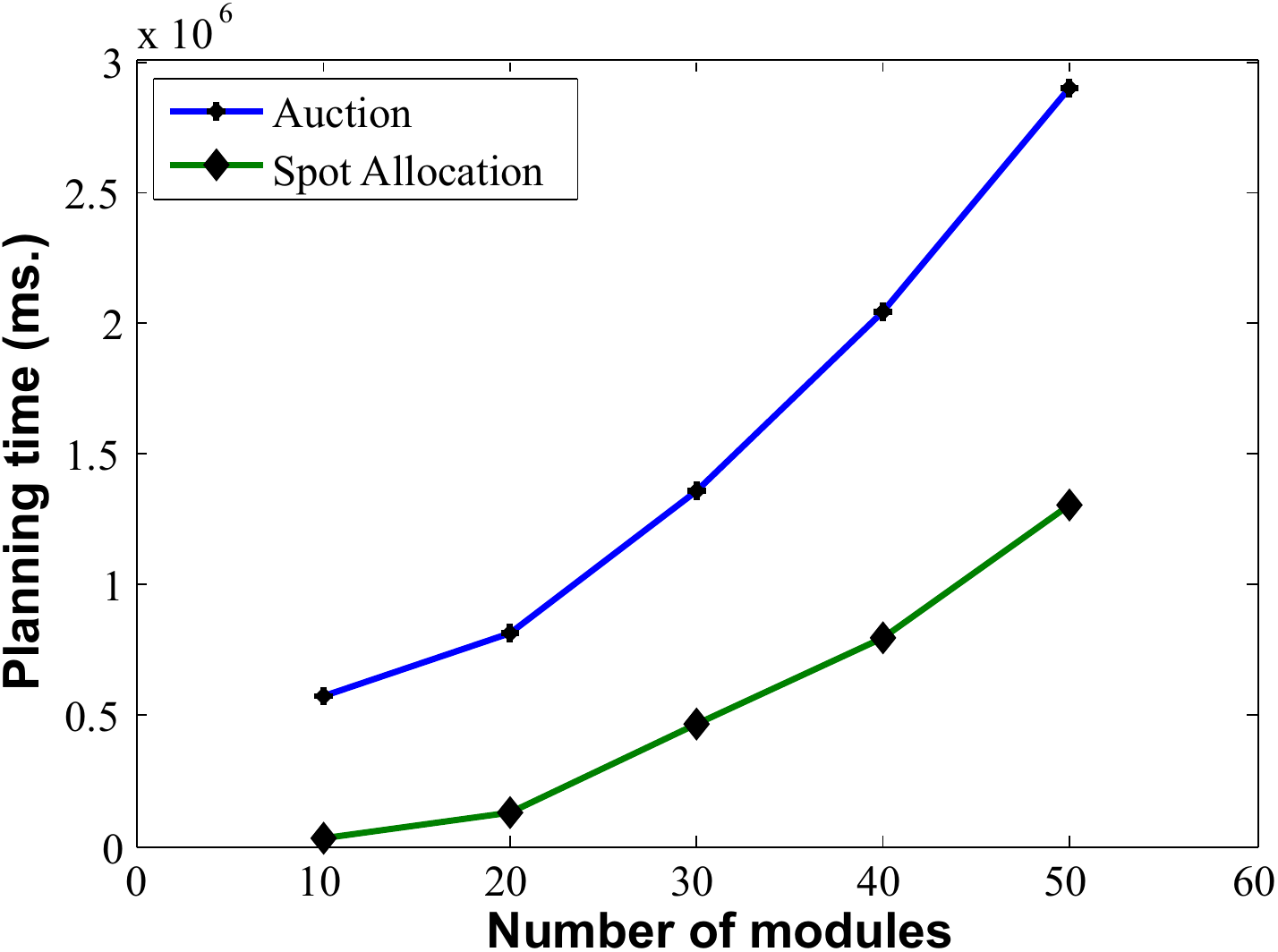}&
\hspace{-0.1in}\includegraphics[width=0.23\linewidth]{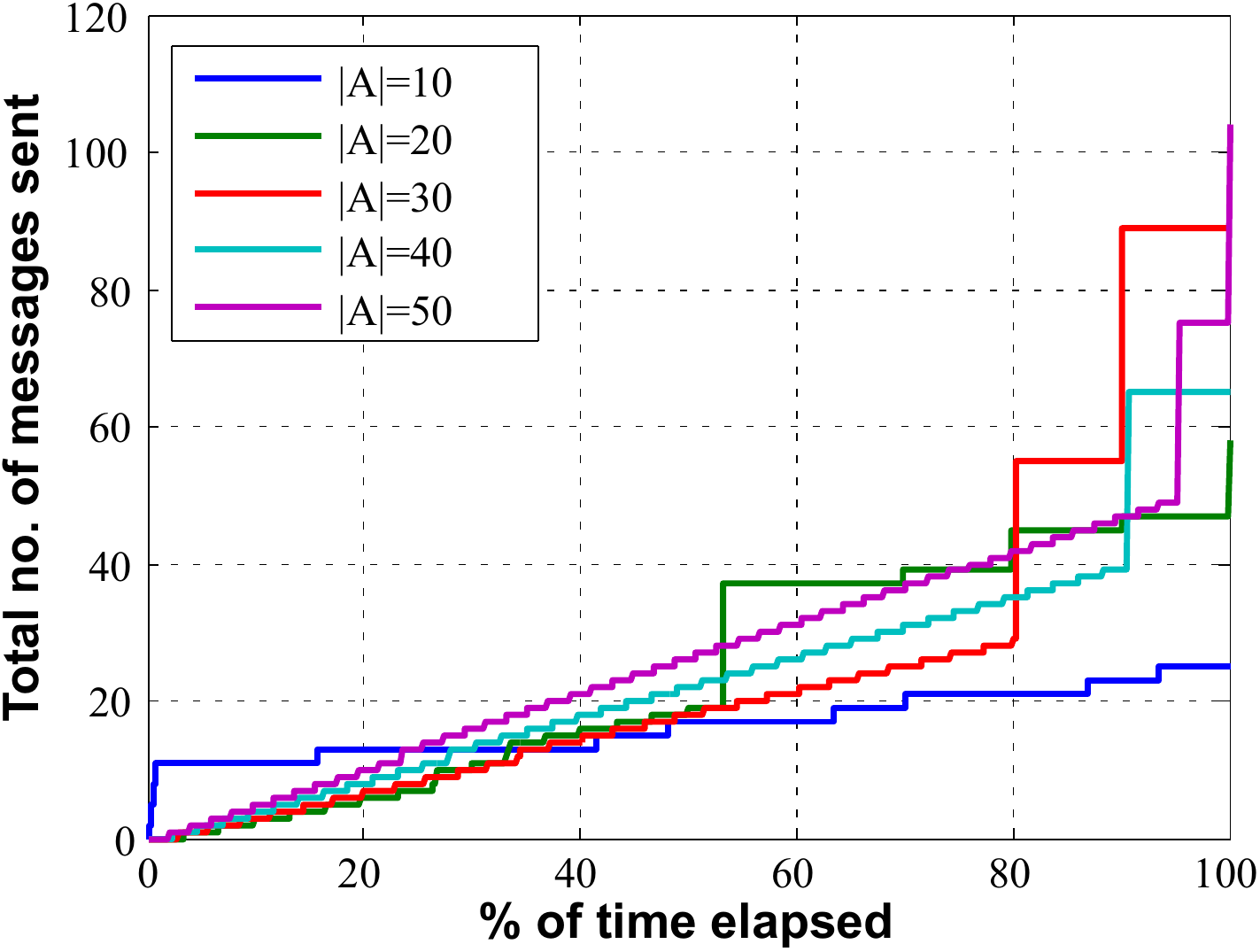}\\
(a) & (b) & (c) & (d)\\
\hspace{-0.3in}\includegraphics[width=0.23\linewidth]{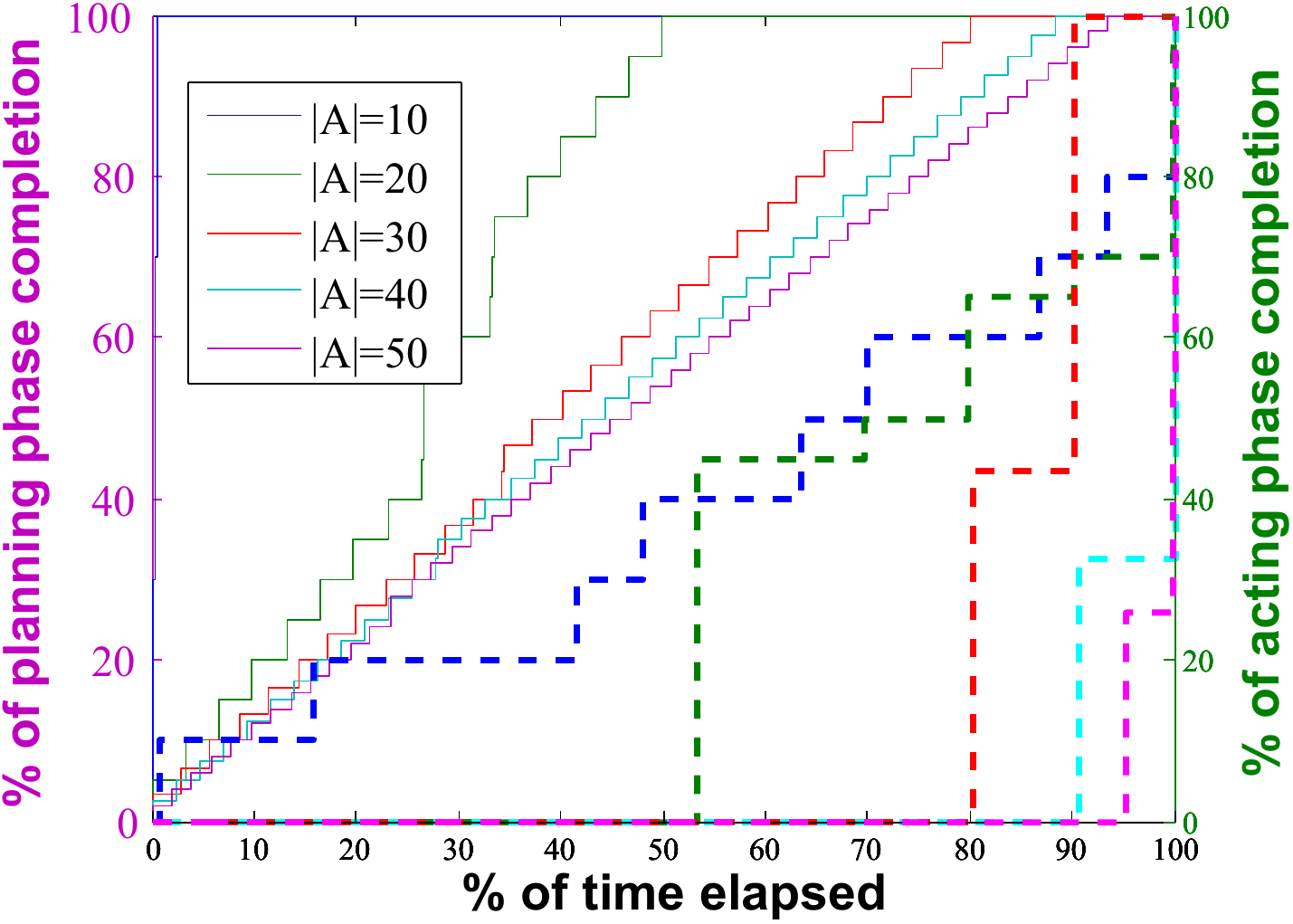}&
\hspace{-0.1in}\includegraphics[width=0.23\linewidth]{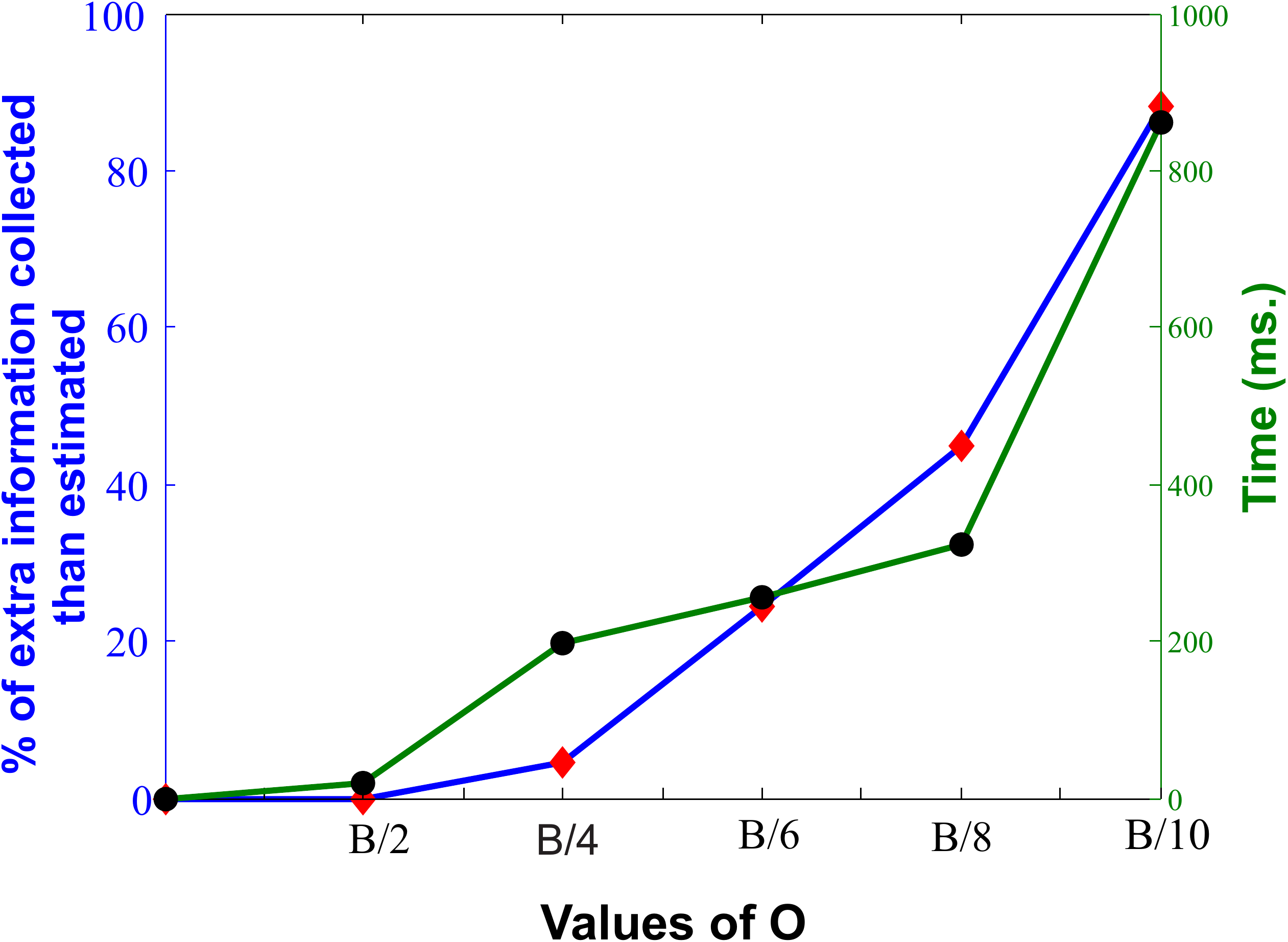}&
\hspace{-0.1in}\includegraphics[width=0.23\linewidth]{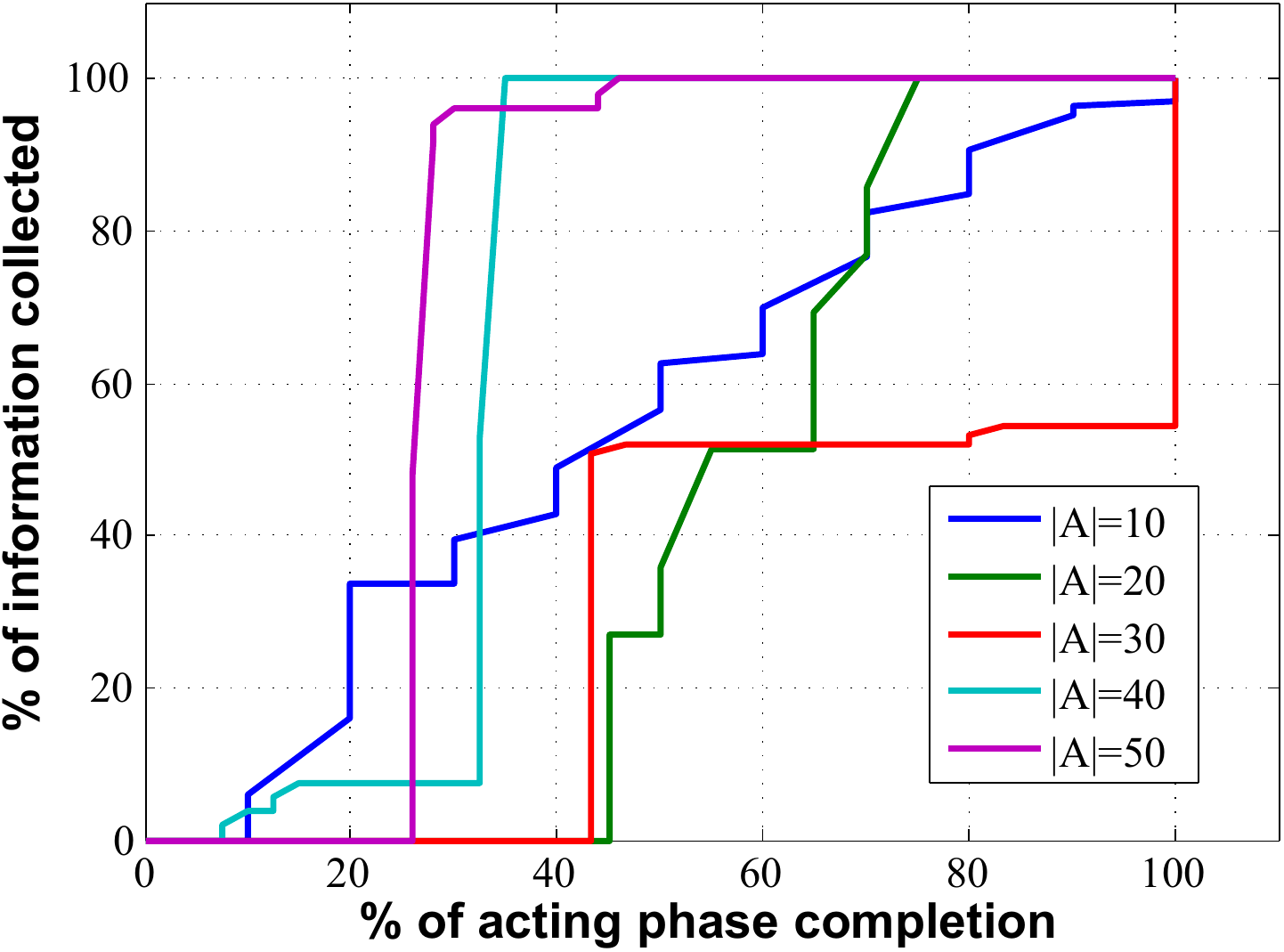}&
\hspace{-0.1in}\includegraphics[width=0.23\linewidth]{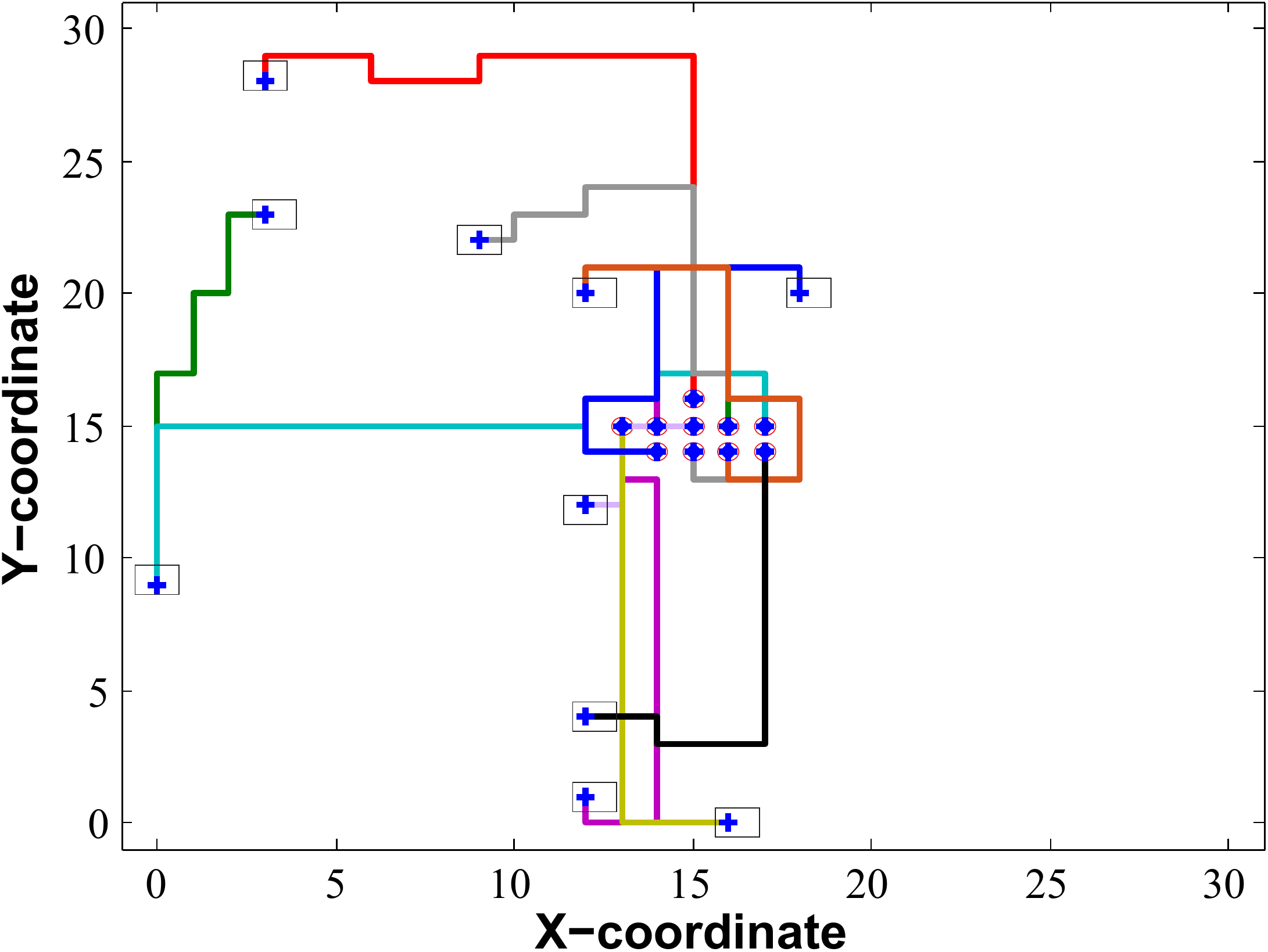}\\
(e) & (f) & (g) & (h)\\
\end{tabular}
\end{center}
\caption{(a) Comparison of estimated information collection between SA and auction algorithm.; (b) Comparison of no. of messages sent in planning phase with auction algorithm; (c) Comparison of planning times between SA and auction algorithm; (d) Message sending rate over time; (e) Planning (solid lines) and acting (dashed lines) phases completion rates; (f) Effect of changing values of $\mathcal{O}$; (g) Change in collected information over time; (h) Configuration formation by $10$ modules: boxed $+$ and circled $*$ indicate the start and final locations respectively.}
\label{all}
\end{figure*}

{\bf Experimental Results.} First, we have tested the run times of the EPS algorithm for different budget amounts. For fixed start and goal locations, $B$ is varied through $[45,50,55]$, where $\mathcal{MD}$(start, goal) $> B$. The result is shown in Figure \ref{timetable}.(a). We can see that with increasing amount of budget, the run time also increases, as the algorithm needs to search for more possible paths in the search space. Figure \ref{timetable}.(b) and (c) show the cells explored by the EPS algorithm for $B = 45$ and $55$ respectively in a particular instance. We have observed that, with $B=55$, on an average the EPS algorithm expanded about $50\%$ more cells in the environment than with $B=45$, which also can be noticed in Figure \ref{timetable}.(b) and (c). 
Next, we compared the performances of proposed SA algorithm and the auction algorithm. In terms of estimated information collection, both the allocation algorithms performed almost equally (Figure \ref{all}.(a)). 
In terms of total number of messages sent by the modules in the planning phase, the SA algorithm outperformed the auction algorithm (Figure \ref{all}.(b)). For $50$ modules, using the auction algorithm, modules have sent about $10^4$ times more number of messages. Figure \ref{all}.(c) shows that auction algorithms takes significantly higher time (with $50$ modules, the auction algorithm takes $3$ times more) than the proposed SA algorithm.

We have observed that the total number of messages sent by the modules increases linearly with the number of modules in the environment. Figure \ref{all}.(d) shows how the count of total messages sent changes for different number of modules over time. We can observe that with increasing number of modules, the rate at which the count of sent messages increases over time, also gets faster. 
Figure \ref{all}.(e) shows the planning and acting phases completion rates for different number of module. $x$-axis denotes the percentage of total time elapsed and two $y$-axes denote how many spots in the target configuration have been allocated to unique modules by the supervisor so far, i.e., percentage of planning completion and how many modules have occupied their spots, i.e., percentage of acting phase completion. We observe that with increasing number of modules involved, more amount of planning time is required. For example, with $20$ modules, planning phase took about $50\%$, whereas for $50$ modules, planning phase took about $90\%$ of total time. For this reason, acting phase amounts also varied largely. This shows that as each module takes more or less the same time to reach the goal spot, the main reason behind the variation in the run times for different number of modules, is the time consumed in the planning phase.

Next, we have varied the value of $\mathcal{O}$ between $\frac{B}{2}$ and $\frac{B}{10}$ to evaluate the effect of frequency of path updates on the information gain and time taken to run the algorithm. This test has been performed with $1$ module only. Result is shown in Figure \ref{all}.(f). We observe that although with increasing number of path updates, the module earned up to $88\%$ extra information than estimated, the running time also increased considerably. For example, with $\mathcal{O} = \frac{B}{2}$, run time is $20$ ms., whereas with $\mathcal{O} = \frac{B}{10}$, run time increased to $860$ ms. In Figure \ref{all}.(g), we have shown how with acting phase completion, the percentage of total information collected by the modules changes. Finally, Figure \ref{all}.(h) shows an instance of the configuration formation procedure. In this experiment, $10$ modules start from arbitrary locations in the environment (boxed $+$ marked points) and form the target configuration, by following the maximally possible informative paths from their initial locations to the allocated goal spots in the target configuration (circled $*$ marked points).

\section{Conclusion and Future Work}
In this paper, we have addressed the problem of simultaneous configuration formation and information collection by modular robots. Our solution uses a centralized sequential allocation technique which allocates the spots in a target configuration to the modules, depending on the estimated amount of information collected by the modules for going to each spot. Our informative path generation technique uses a best-first search to find a path within the given budget. In the future, we plan to extend this algorithm to move the modules in parallel instead of our current sequential movement strategy which will reduce the time for acting phase. We also plan to extend this algorithm to avoid overlaps in robots' paths and in effect avoid redundant information collection. We are also planning to implement this algorithm on physical ModRED hardware.

{\small 
\bibliographystyle{abbrv}
\bibliography{references}
}
\end{document}